\documentclass[journal]{IEEEtran}
\usepackage{subfigure}
\usepackage{epsfig, latexsym, amssymb, amsmath}
\usepackage{amsthm}
\usepackage{graphics}
\usepackage{bm}
\usepackage{cite}
\usepackage{graphicx}
\usepackage[mathscr]{eucal}
\usepackage{color}
\usepackage{dsfont}
\usepackage{url}
\usepackage{dirtytalk}
\usepackage{caption}

\usepackage{enumitem}
\usepackage{algorithm}
\usepackage{algorithmicx}
\usepackage{hyperref}

\newcommand{\revv}[1]{{\color{black}#1}} 

\usepackage[utf8]{inputenc}

\begin{document}
\title{Robust Event Classification Using Imperfect Real-world PMU Data
}

\author{
        Yunchuan Liu,~\IEEEmembership{Graduate Student Member,~IEEE,}  Lei~Yang,~\IEEEmembership{Senior Member,~IEEE,}
        Amir Ghasemkhani,~\IEEEmembership{Member,~IEEE,}
        Hanif Livani,~\IEEEmembership{Senior Member, IEEE,}
        Virgilio A. Centeno,~\IEEEmembership{Senior Member, IEEE,}
        Pin-Yu Chen,~\IEEEmembership{Member, IEEE,}
        Junshan Zhang,~\IEEEmembership{Fellow, IEEE}
\thanks{
Yunchuan Liu and Lei Yang are with the Department of Computer Science and Engineering, University of Nevada Reno, Reno,
NV, 89557 USA (e-mail: ~ycliu@nevada.unr.edu, leiy@unr.edu).

 Amir Ghasemkhani is with the Computer Science and Engineering, California State University San Bernardino, San Bernardino, CA,92407 USA.(e-mail: ~amir.ghasemkhani@csusb.edu).
 
Hanif Livani is with the Department of Electrical and Biomedical Engineering, University of Nevada Reno, Reno, NV 89557 USA (e-mail:~ hlivani@unr.edu).

Virgilio A. Centeno is with the Department of Electrical and Computer Engineering, Virginia Polytechnic Institute and State University, Blacksburg, VA 24061 USA (e-mail:~ virgilio@vt.edu).

 Pin-Yu Chen is with the IBM Thomas
J. Watson Research Center, Yorktown Heights, NY 10598 USA (e-mail:~pin-yu.chen@ibm.com).

Junshan Zhang is with the School of Electrical,
Computer and Energy Engineering, Arizona State University, Tempe, AZ 85287 USA (e-mail:~Junshan.Zhang@asu.edu).

}

        }

\date{July 2021}

\maketitle
\begin{abstract}
This paper studies robust event classification using imperfect real-world phasor measurement unit (PMU) data. By analyzing the real-world PMU data, we find it is challenging to directly use this dataset for event classifiers due to the low data quality observed in PMU measurements and event logs. To address these challenges, we develop a novel machine learning framework for training robust event classifiers, which consists of three main steps: data preprocessing, fine-grained event data extraction, and feature engineering. Specifically, the data preprocessing step addresses the data quality issues of PMU measurements (e.g., bad data and missing data); in the fine-grained event data extraction step, a model-free event detection method is developed to accurately localize the events from the inaccurate event timestamps in the event logs; and the feature engineering step constructs the event features based on the patterns of different event types, in order to improve the performance and the interpretability of the event classifiers. Based on the proposed framework, we develop a workflow for event  classification using the real-world PMU data streaming into the system in real time. Using the proposed framework, robust event classifiers can be efficiently trained based on many off-the-shelf lightweight machine learning models. Numerical experiments using the real-world dataset from the Western Interconnection of the U.S power transmission grid show that the event classifiers trained under the proposed framework can achieve high classification accuracy while being robust against low-quality data.


\end{abstract}


\section{Introduction}
\subsection{Motivation and Related Works}
Recent years have witnessed the booming deployment of phasor measurement units (PMUs) \cite{naspi}. More than 2500 PMUs are installed in the North American power system. Compared to traditional supervisory control and data acquisition (SCADA) systems, PMUs are of much higher
sampling rates (e.g., 30 or 60 samples per second in the U.S.), which provides golden opportunities to achieve high level of situational awareness (e.g., real-time event detection and classification), in order to prevent large-scale blackouts (e.g., \cite{white20032003,2013storm}). In this paper, we study robust event classification using real PMU data.

Much effort has been made on the development of PMU based event detection and classification. 
For PMU based event detection, many methods have been developed (e.g., \cite{negi2017event,senaratne2021spatio,cui2018novel,liu2019data,zhou2018nonparametric,xie2014dimensionality,ling2019new,shi2020online}).
Compared to PMU based event detection, large amounts of high-quality labeled PMU datasets are critical for the development of PMU based event classification, especially for the development of neural network based classifiers. However, based on our observations from large amounts of real\revv{-world} PMU data (see Section \ref{sec:data-description}), high-quality labeled PMU datasets are not available in practice.
In fact, many existing works on PMU based event classification consider a small amount of PMU data with a few labeled events. For example, in \cite{li2018real}, the dataset consists of only 32 labeled events; in \cite{dahal2012preliminary}, only 4 PMUs are used in case studies; in \cite{nguyen2015smart}, only 57 labeled line events are used to train an event classifier; in \cite{wang2020frequency}, hundreds of labeled frequency events from the FNET/GridEye system are used to train a Convolutional Neural Network (CNN) based frequency event detector. The generalization of event classifiers trained using a small dataset can be poor.

To address the challenge of insufficient PMU data, some studies leverage synthetic data generated by simulation or neural networks. For instance, in \cite{liu2019data}, simulated data with man-made noises are used; 
in \cite{zheng2021generative}, Generative Adversarial Networks (GAN) are used to generate synthetic event data. While synthetic data can increase the amounts of data for training event classifiers, the events and grid characteristics hidden in the real PMU data can hardly be represented by synthetic data. Thus, the generalization of event classifiers trained using synthetic data can still be poor.



This paper leverages large amounts of real\revv{-world} PMU data from the Western Interconnection of continental U.S. transmission grid (see the data description in Section \ref{sec:data}) for the development of event classifiers. By analyzing this large dataset, we find that it is challenging to directly use this dataset to train event classifiers due to the low data quality. Specifically, the real PMU data are noisy and contain bad data, dropouts, and timestamp errors. The  timestamps of labeled events provided in the event logs are inaccurate. Though the entire dataset contains measurements from many PMUs over a two-year period, the total number of labeled events is only a few thousands and the distribution of different event types is highly imbalanced. Due to these issues, the performance of off-the-shelf machine learning models directly trained using such dataset can be significantly degraded. 

Recently, there have been several attempts using large-scale real PMU data to develop event classifiers based on neural networks (e.g., auto-encoder model \cite{leao2020big},  Convolutional Neural Network (CNN) \cite{niazazari2021pmu}, spatial pyramid pooling (SPP)-aided CNN \cite{yuan2021learning}, and information loading enhanced Deep Neural Network (DNN) \cite{shi2021power}). However, the hyperparameter tuning of the neural network based approaches is challenging and the training time is long. Though good classification results using neural networks are reported \cite{leao2020big,niazazari2021pmu,yuan2021learning,shi2021power}, the interpretability of neural networks is low, not to mention the adversarial vulnerability of neural networks.

\subsection{Main Contributions}
To address these challenges, this paper develops a novel machine learning framework for training event classifiers using large-scale imperfect real\revv{-world} PMU data, which consists of three main \revv{steps}: data preprocessing, fine-grained event data extraction, and feature engineering. The goal is to obtain high-quality labeled PMU training data from the low-quality real PMU data. Specifically, the data preprocessing \revv{step} addresses the data quality issues of PMU measurements (e.g., bad data and missing data); the fine-grained event data extraction \revv{step} accurately localizes the events from the inaccurate event timestamps in the event logs by using a model-free event detection method developed based on the low-rank property of PMU data; and the feature engineering \revv{step} constructs the event features based on the patterns of different event types, in order to improve the performance and the interpretability of the event classifiers. Based on the proposed machine learning framework, we also develop a workflow for event classification using the real PMU data streaming into the system in real time.
One salient merit of the proposed framework is that large-scale real PMU data is reduced to a small set of event features, which can be used to efficiently train many off-the-shelf lightweight machine learning models. As the features are constructed based on the event patterns, the contribution of any single PMU measurement to the features is low, which can effectively mitigate the impact of bad and missing data and improve the robustness of the event classifiers.


Using the two-year real-world PMU data from the Western Interconnection of continental U.S. transmission grid, we evaluate the performance of the proposed machine learning framework. Many off-the-shelf lightweight machine learning models can be efficiently trained in our framework and achieve good performance. For example, the training time of the Random Forest model is less than 2 seconds while the testing accuracy of the Random Forest model is 94\%. Moreover, our framework demonstrates a strong robustness against missing data. Even under the missing rate of 50\%, the accuracy of the Random Forest model can still achieve 87\%. In summary, the proposed machine learning framework provides a promising way to train robust event classifiers with good interpretability and low training cost.



The rest of the paper is organized as follows: Section II describes the real-world PMU data used in this paper and elaborates the key challenges of using real PMU data for the development of event classifiers. Section III provides the details of the proposed machine learning framework. Section IV evaluates the performance of event classifiers trained under the proposed framework. Section V concludes the paper.

\section{Data Description and Key challenges}\label{sec:data-description}
In this section, we introduce the real-world PMU data used in this study and identify the key challenges of using this dataset for the development of event classifiers.

\subsection{Data Description}
\label{sec:data}
This paper uses real-world PMU data from the Western Interconnection of continental U.S. transmission grid. The dataset is complied by the Pacific Northwest National Laboratory (PNNL) to anonymize the data such that proprietary information (e.g., PMU locations,  event locations, and the system topology) is unavailable.
The dataset contains measurements from 43 PMUs over a two-year period (2016–2017). The sampling rates of PMUs are either 30 or 60 frames per second. 
The size of the dataset is about 5 TB (stored in Parquet format), which contains over 93 billion records. In each record, the measurements contain: 1) coordinated universal time (UTC), 2) voltage magnitude of positive sequence, A phase, B phase, and C phase, 3) voltage angle of positive sequence, A phase, B phase, and C phase, 4) current magnitude of positive sequence, A phase, B phase, and C phase, 5) current angle of positive sequence, A phase, B phase, and C phase, 6) frequency, 7) rate of change of frequency (ROCOF), 8) PMU status flag, and 9) anonymized PMU ID.
 

Besides the raw PMU measurements, event logs are provided for the development of event classifiers. In the event logs, four types of events (i.e., line outage, transformer outage, frequency event, and oscillation event) are recorded. For each event, start timestamp, end timestamp, event type, event cause, and event description are provided. The total number of events in the event logs is 4,854, including
3,667 line outages, 621 transformer outages, 465 frequency events, and 100 oscillation events.






\subsection{Key Challenges}

By analyzing the real-world PMU measurements and the event logs, we find that it is challenging to directly use this dataset to develop event classifiers as the data quality is low and off-the-shelf machine learning approaches require high-quality training data. Specifically, we face the following major challenges of using this dataset for the development of event classifiers.

\begin{figure}[!t]
\centering
\includegraphics[scale=0.36]{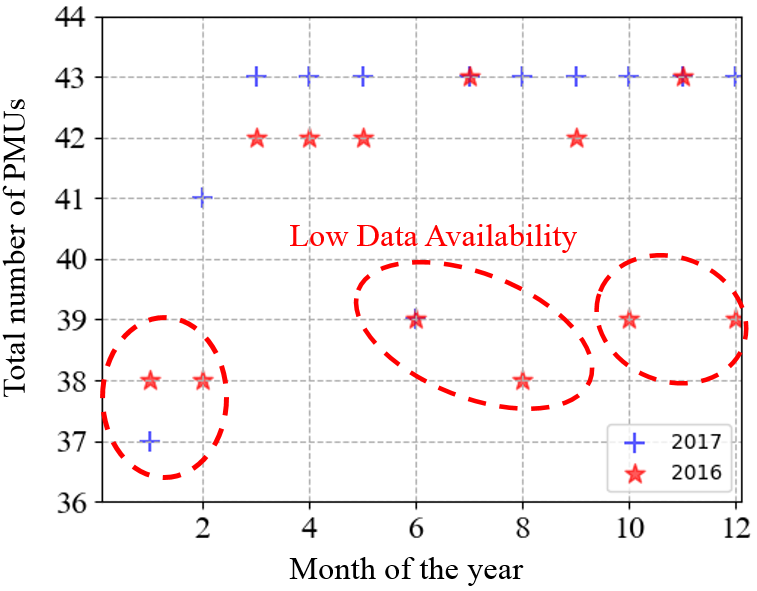}
\caption{PMU data availability.}
\label{fig:ava}
\end{figure}

\begin{figure}[!t]
\centering
\includegraphics[scale=0.52]{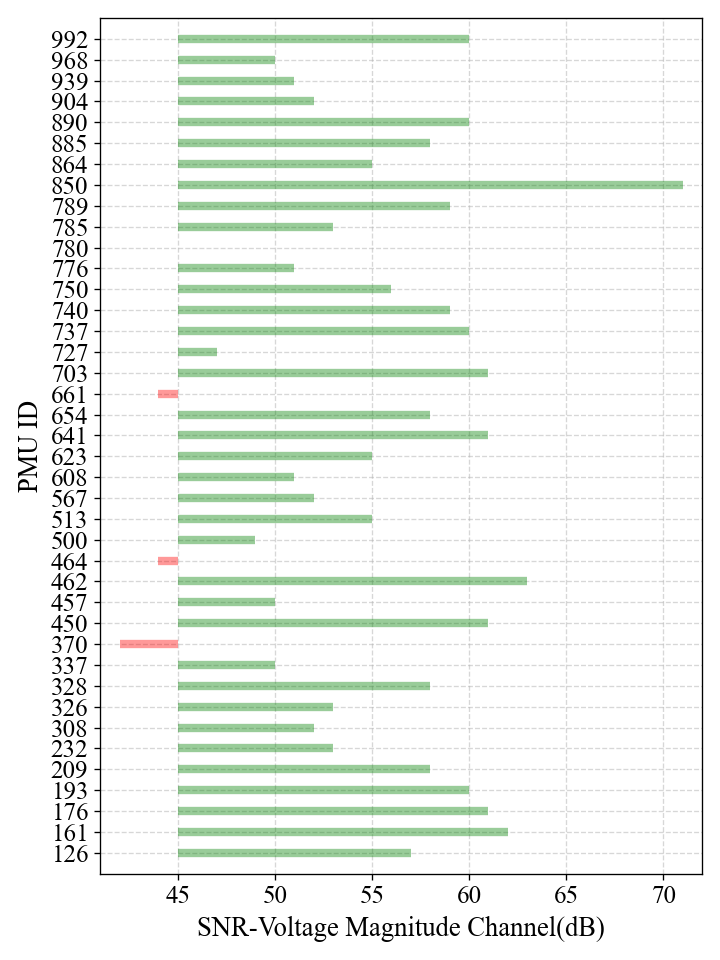}
\caption{SNR of voltage magnitude of positive sequence of different PMUs.}
\label{fig:snr}
\end{figure}

\subsubsection{Incomplete and noisy PMU measurements}
Fig. \ref{fig:ava} illustrates the availability of PMU data, where the data from some PMUs are completely missing in certain months. Moreover, the current magnitude of A phase, B phase and C phase and the current angle of A phase, B phase and C phase are unavailable in most cases. 
Fig. \ref{fig:snr} illustrates the signal-to-noise ratio (SNR) of PMUs. Based on \cite{7741972}, 45 dB can be used as a threshold to indicate whether the PMU measures are noisy. As shown in Fig. \ref{fig:snr}, three PMUs are below 45 dB. Therefore, the quality of PMU measurements needs to be accounted for when preparing for the training dataset.

\subsubsection{Inaccurate event timestamps in the event logs}
\begin{figure*}[th]
	\begin{minipage}[t]{0.33\linewidth}
		\centering
		\includegraphics[width=2.2in]{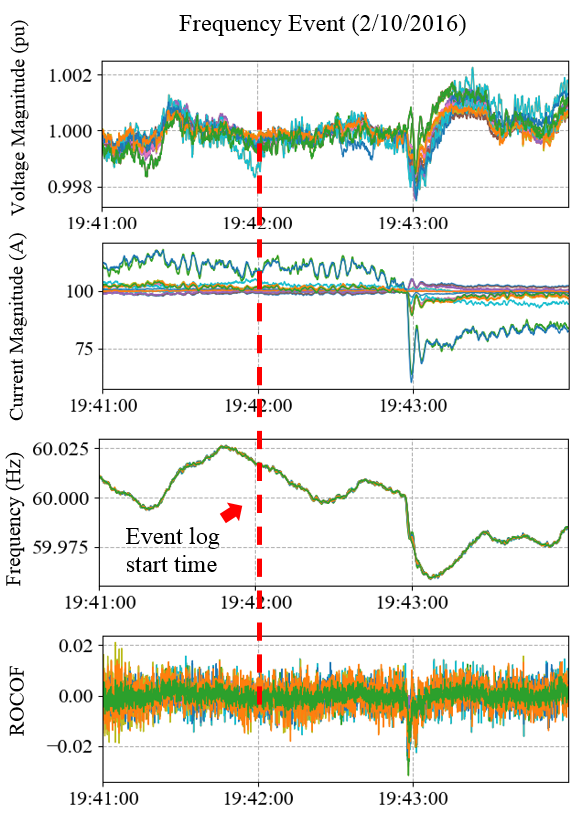}
		\caption{Example of a frequency event.}
		\label{fig:freq}
	\end{minipage}
	\begin{minipage}[t]{0.33\linewidth}
		\centering
		\includegraphics[width=2.2in]{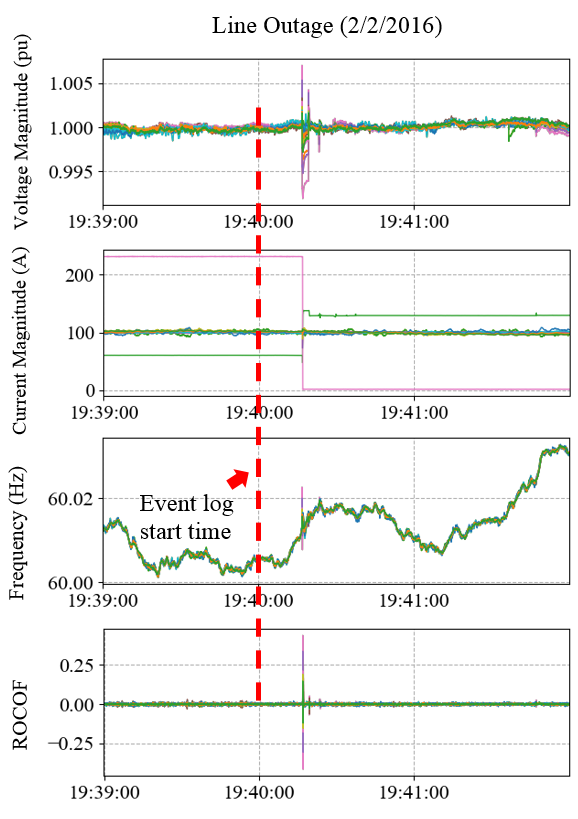}
		\caption{Example of a line outage.}
		\label{fig:line}
	\end{minipage}
	\begin{minipage}[t]{0.33\linewidth}
		\centering
		\includegraphics[width=2.2in]{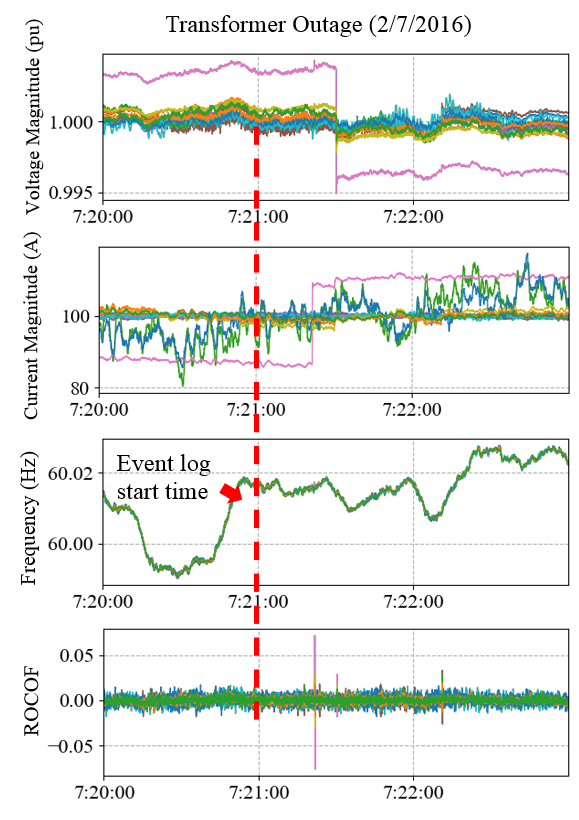}
		\caption{Example of a transformer outage.}
		\label{fig:trans}
	\end{minipage}
	\vspace{-0.2cm}
\end{figure*}

By analyzing the event logs, we observe that the timestamps of events provided in the event logs are inaccurate. Figs. \ref{fig:freq}--\ref{fig:trans} illustrate the PMU measurements during different events, where the red vertical line indicates the start time of the event provided in the event logs. For example, in Fig. \ref{fig:freq}, the actual start time of the frequency event is around 19:43:00, while the start time provided in the event logs is 19:42:00. 
Clearly, the timestamps in the event logs are not accurate. If such timestamps are used for event extraction, with high probability, the event features would be missing and therefore the performance of event classification would be degraded.


\subsubsection{Insufficient and imbalanced training data}
The total number of events in the event logs is only a few thousands, and 75\% of the events are line outages. Directly applying off-the-shelf machine learning models on such dataset can lead to overfitting issues, especially for neural network based models, where the parameters of neural networks can be much larger than the number of events in the training dataset. To address this challenge, recent works on neural networks (e.g., \cite{yuan2021learning,shi2021power}) leverage data augmentation and report good classification results; however, the training time of neural networks is long and the interpretability of neural networks is low. As shown in the next section, different types of the events exhibit distinct features, based on which good machine learning models with better interpretability and low training cost can be built.

\section{A Machine Learning Framework for Robust Event Classification}


\subsection{The Machine Learning Framework for Real-world PMU Data}

\begin{figure}[t]
\centering
\includegraphics[scale=0.55]{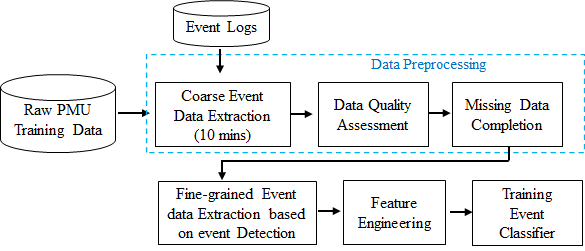}
\caption{The machine learning framework for robust event classification.}
\label{fig:train}
\end{figure}

To address the challenges of using the real-world PMU data for the development of event classifiers, we propose a novel machine learning framework to handle these challenges, as illustrated in Fig. \ref{fig:train}. 
Specifically, we first do data preprocessing: 1) we coarsely localize the events based on the inaccurate event timestamps in the event logs, where a large time window (e.g., 10 minutes) around the event timestamp in the event logs is used to ensure that the event features are included; 2) we then do data quality assessment to filter out bad readings, which are treated as ``missing'' data; and 3) we complete all the missing data based on our prior work \cite{ghasemkhani2020regularized}. Then, we do fine-grained event data extraction to accurately localize the events using event detection, where a much smaller time window (e.g., 5 seconds) will be used to ensure that only the event data is included. Based on the fine-grained event data extraction, more distinct features can be constructed for each event type. These features will be used to train event classifiers.
In the following, we  discuss the details of each component in the proposed machine learning framework.

\subsection{Data Preprocessing}\label{sec:preprocessing}
As the event timestamps in the event logs are inaccurate, we extract the data in a large time window (i.e., 10 minutes) centered at the start time of an event provided in the event logs (i.e, 5 minutes before and 5 minutes after the start time), in order to preserve the event features. In the raw PMU dataset, the measurements of all PMUs in each day are stored in one Parquet file and these measurements are not ordered according to the timestamps. When doing the event data extraction, we  partition the data based on PMU ID and sort the corresponding measurements based on the timestamps. As mentioned in Section \ref{sec:data-description}, certain signals (e.g., the current magnitude of A, B and C phases) are missing in most cases. In the event data extraction, we extract the voltage magnitude and the current magnitude of positive sequence and ROCOF for the development of event classifiers.



Then, we do data quality assessment to filter out bad data. As the PMU status flags are provided, we first remove the measurements when PMUs were malfunctioning or in test mode. After the status check, we further remove the bad data based on the following criteria:
\begin{itemize}
\item Remove the measurements out of physical bound, i.e., the angle measurement larger than 180 or less than zero and the current or voltage magnitude less than zero.

\item Remove the measurements that are 3 times greater than the standard deviation based on the empirical distribution of the measurements.

\end{itemize}
The bad data are treated as ``missing'' data, which are completed based on our regularized tensor completion  approach \cite{ghasemkhani2020regularized}.






\subsection{Event Detection based Fine-grained Data Extraction} \label{sec:detect}

\begin{figure}[!ht]
\centering
\includegraphics[scale=0.43]{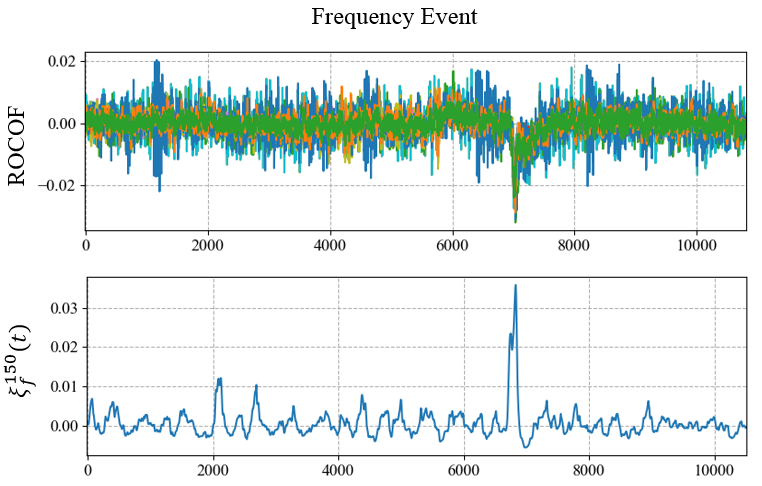}
\caption{Frequency event detection based on the low-rankness of PMU data, where $\xi_f^{150}(t)$ of the ROCOF signal is shown in the bottom and $w=150$ is used.}
\label{fig:locate}
\end{figure}

The goal of fine-grained data extraction is to accurately extract the event data for better constructing event features, as the event timestamps provided in the event logs are not accurate. To this end, we develop a model-free event detection method based on the low-rank property of PMU data to accurately localize the events. We observe that when the disturbance occurs in the system, the low-rankness of PMU data will change (see Fig. \ref{fig:locate}), which can be quantified using the singular values of PMU measurement matrices.

Specifically, let $M^w_{s}(t) \in \mathbb{C}^{w \times n}$ be a PMU measurement matrix for an extracted signal $s\in\mathcal{S}$, which embraces the past $w$ measurements before the timestamp $t$ from $n$ PMUs. Here $\mathcal{S}$ denotes the set of extracted signals including the  voltage  magnitude  and  the  current  magnitude  of  positive sequence and ROCOF. For $M^w_{s}(t)$, we do the singular value decomposition (SVD) and compute the ratio of the largest $\sigma_1$ and the second largest singular $\sigma_2$ values, i.e., $\eta_t = \frac{\sigma_2}{\sigma_1}$. Then, the average relative change of this ratio in the time window $w$ is calculated as
\begin{align}
\xi_s^w(t)  = \frac{\eta_t - \eta_{t-w} }{\eta_{t-w}\cdot w}.
\label{eq:re_change}
\end{align}
Based on these $\xi_s^w(t)$ from different signals, we use a threshold-based OR rule to determine whether there is an event. In other words, an event is detected, if one of these $\xi_s^w(t)$ is greater than a pre-determined threshold $\theta_s$. Fig. \ref{fig:locate} gives an example of detecting a frequency event using the ROCOF signal, where $\xi_f^w(t)$ changes significantly when the frequency event occurs.
When an event is detected, we extract the event data in a smaller time window $W$ (e.g., 5 seconds) centered at the detected event start time.

The proposed event detection approach is model-free with the parameters of the window size $w$ and the detection thresholds $\bm{\theta}=\{\theta_s\}$ of extracted signals $\mathcal{S}$. To optimize the detection performance, we tune the parameters by a Baysian optimization algorithm, which can efficiently search for the best parameters. The idea of Baysian optimization is to model the unknown function between the parameters and the detection errors using a multivariate  Gaussian  distribution, and then use a computationally cheap acquisition function to guide the search. We introduce  an  acquisition  function $\beta(\cdot)$ as the  optimization  objective,  which  characterizes  the  expected detection error improvement under $(\bm{\theta},w)$:
\begin{equation}
\beta(\bm{\theta},w) = \mathbb{E}[(e(\bm{\theta}^*,w^*) - e(\bm{\theta},w))^+],
\end{equation}
where $e(\bm\theta,w)$ denotes the detection error under $(\bm{\theta},w)$, and $e(\bm\theta^*,w^*)$ denotes the lowest detection error that has been obtained so far. It is assumed that the detection errors are random variables following the multivariate Gaussian distribution: $\mathcal{G} \sim \mathcal{N}(m(\bm\theta,w), Cov(\bm\theta,w))$ with mean $m(\bm\theta,w)$ and covariance $Cov(\bm\theta,w)$.    
In each iteration, we find  $(\bm{\theta},w)$ that maximizes the acquisition function $\beta(\bm\theta,w)$. 
Then, $(\bm{\theta},w)$ and the corresponding $e(\bm{\theta},w)$ will be added into a sample set $\mathcal{S}$, and the mean $m(\bm\theta,w)$ and covariance $Cov(\bm\theta,w)$  of $\mathcal{G}$ will be updated accordingly \cite{snoek2012practical}. The details of the Bayesian optimization based parameter search are given in Algorithm \ref{alg:bo}.

\begin{algorithm}[H]
\caption{Bayesian optimization based parameter search}
\label{alg:bo}
\begin{algorithmic}[0]
\State \textbf{Initialization:} Initialize  $\mathcal{S}=\{((\bm{\theta},w), e(\bm{\theta},w))\}$.
\State \textbf{For each iteration:}
\State 1) Find the parameters $(\bm{\hat \theta},\hat w)$ that maximize $\beta$, i.e.,  $(\bm{\hat \theta},\hat w)= {\arg\max}_{((\bm{\theta},w), e(\bm{\theta},w))\in\mathcal{S}} \beta (\bm{\theta},w)$. 

\State 2) Use $(\bm{\hat \theta},\hat w)$ for event detection and compute the corresponding detection error $e(\bm{\hat \theta},\hat w)$.

\State 3) Add $((\bm{\hat \theta},\hat w),e(\bm{\hat \theta},\hat w))$ into the sample set $ \mathcal{S} = \mathcal{S}\cup((\bm{\hat \theta},\hat w),e(\bm{\hat \theta},\hat w))$, and update the parameters of $m(\bm{\theta},w)$ and $Cov(\bm{\theta},w)$ using $\mathcal{S}$.

\end{algorithmic}
\end{algorithm}



\subsection{Feature Engineering} 
\label{sec:feature}

\begin{figure*}[th]
	\begin{minipage}[t]{0.33\linewidth}
		\centering
		\includegraphics[width=2.2in]{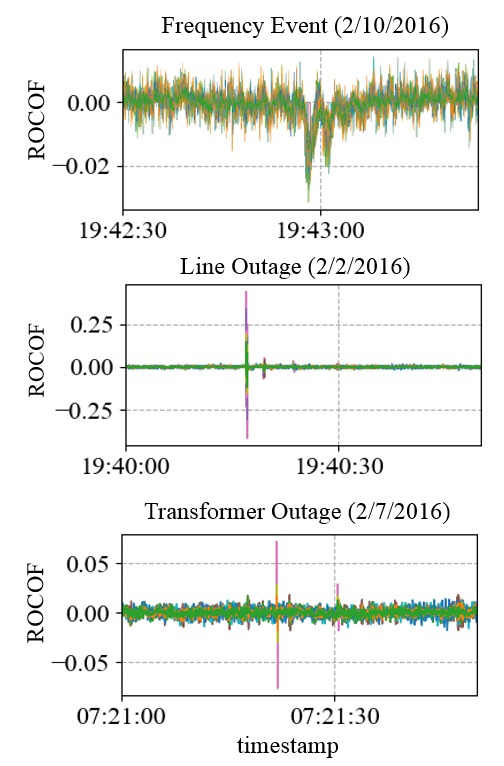}
		\caption{Comparison of ROCOF in different events.}
		\label{fig:rof}
	\end{minipage}
	\begin{minipage}[t]{0.33\linewidth}
		\centering
		\includegraphics[width=2.2in]{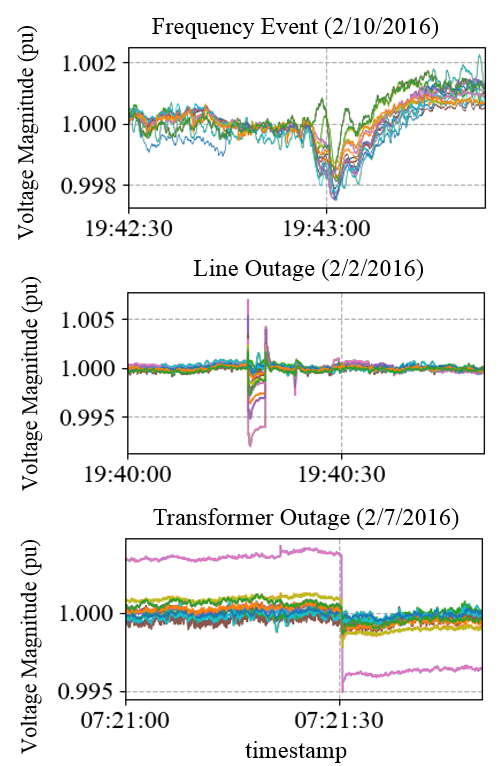}
		\caption{Comparison of voltage magnitude in different events.}
		\label{fig:vpm}
	\end{minipage}
	\begin{minipage}[t]{0.33\linewidth}
		\centering
		\includegraphics[width=2.2in]{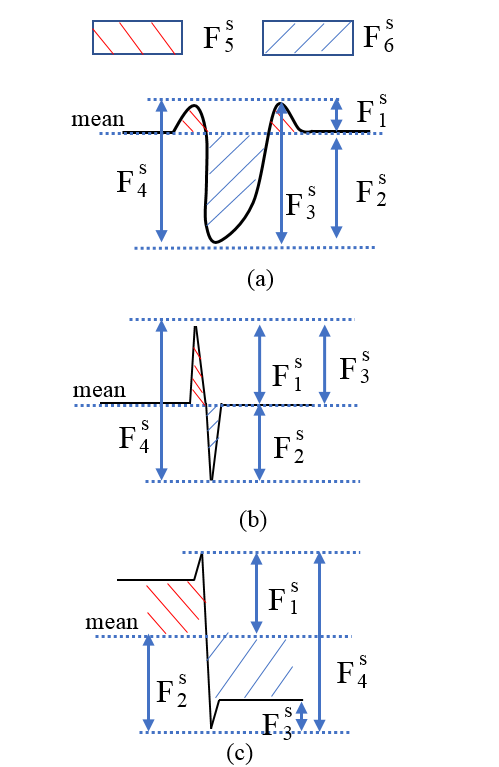}
		\caption{Illustration of constructed features for different event types.}
		\label{fig:shiyi}
	\end{minipage}
	\vspace{-0.2cm}
\end{figure*}

Using the event data from fine-grained data extraction, we construct the event features based on the patterns of different event types. By carefully analyzing the PMU data during events, we find that the shape and the duration of the signals under different event types are distinct:
\begin{itemize}
    \item The ROCOF measurements of frequency events tend to have a deeper and wider dip, compared to a narrow spark of line and transformer outages (see Fig. \ref{fig:rof}).
    
    \item The voltage magnitude measurements of transformer outages tend to have a cliff-like drop with a longer duration time, compared with a narrow spark of line outages (see Fig. \ref{fig:vpm}). 
    
    \item The signal similarity among different PMUs under different event types is different. 
    Frequency events can be observed by all PMUs, while only a few PMUs can capture line and transformer outages.  
\end{itemize}
Based on these observations, we construct the event features.

Specifically, let $X^s_{i}(t)$ denote the measurement of the $i$th PMU's signal $s\in\mathcal{S}$ at time $t$. 
In the event detection based fine-grained data extraction, we extract the event data in a small time window $W$. Let $\mathcal{W}$ denote the set of timestamps in this time window $W$.
Using the measurements $\{X^s_{i}(t), t\in\mathcal{W}\}$ of each PMU's signal in this window, we construct the following six event features as illustrated in Fig. \ref{fig:shiyi}:
\begin{itemize}
    \item Amplitude above the average: 
    \begin{align}
    F^{s}_{1}(i) = \max_{t \in \mathcal{W}}\{X^s_{i}(t) - \overline {X^s_i}\}
     \label{eq:up}
    \end{align}
    \item Amplitude below the average:
    \begin{align}
    F^{s}_{2}(i) = \max_{t \in \mathcal{W}}\{\overline {X^s_i} - X^s_{i}(t)\} 
    \end{align}   
    \item Ramp-up rate:
    \begin{align}
    F^s_{3}(i) = \max_{t_1>t_2, t_1,t_2\in \mathcal{W}}\{X^s_{i}(t_1) - X^s_{i}(t_2)\}
    \end{align}
    \item Ramp-down rate:
    \begin{align}
    F^s_{4}(i) = \max_{t_1<t_2, t_1,t_2\in \mathcal{W}}\{X^s_{i}(t_1) - X^s_{i}(t_2)\}
    \end{align}
    \item Area above the average:
    \begin{align}
        F^s_{5}(i) = \sum_{\{t|X^s_{i}(t)>\overline {X^s_i}, t\in\mathcal{W}\}}  (X^s_{i}(t)-\overline {X^s_i})
    \end{align}
    \item Area below the average:
    \begin{align}
        F^s_{6}(i) = \sum_{\{t|X^s_{i}(t)<\overline {X^s_i}, t\in\mathcal{W}\}}  (\overline {X^s_i}-X^s_{i}(t))
         \label{eq:dn_area}
    \end{align}
\end{itemize}
where $\overline {X^s_i}$ is the average of the $i$th PMU's signal $s$, i.e., $\overline{X^s_i} = \frac{1}{W} \sum_{t\in\mathcal{W}} X^s_i(t)$. 
These features aim to capture the shape features of different event types shown in Figs. \ref{fig:rof} and \ref{fig:vpm}. Fig. \ref{fig:shiyi} illustrates the constructed features for each event type, where
Fig. \ref{fig:shiyi}(a) corresponds to the typical shape of ROCOF in a frequency event, Fig. \ref{fig:shiyi}(b) corresponds to the typical shape of voltage/current magnitude in a transformer outage, and Fig. \ref{fig:shiyi}(c) corresponds to the typical shape of ROCOF in a transformer/line outage.

To capture the signal similarity among different PMUs, we compute the maximum $F_j^{s,max}$, the minimum $F_j^{s,min}$, and the mean $\overline{F^s_{j}}$ values of each feature $j=1,...,6$ based on $n$ PMUs' features $\{F_j^s(i), i=1,...,n\}$. As the voltage magnitude, the current magnitude, and ROCOF are considered in the extracted event data, the number of features constructed for each event is 54 ($6 \times3\times3$).


Based on our experiments, we find that the auxiliary ratio features $\{\overline{F^s_{1}}/\overline{F^s_{2}}, \overline{F^s_{5}}/\overline{F^s_{6}},\overline{F^s_{5}}/(\overline{F^s_{5}}+\overline{F^s_{6}})\}$ for ROCOF constructed based on $\overline{F^s_{j}}$ can better capture the signal shape. Therefore, in our experiments, we use these 57 (54+3) features of each event to build event classifiers.

\textbf{Remarks.} The proposed features are constructed based on the patterns of different event types observed in real-world PMU data. Compared to neural network based approaches, the number of features in our approach is much less than the large number of automatically generated features (e.g., CNN), and the interpretability of our approach is much better. Moreover, with the small number of good features, we need much less number of training data to train a good event classifier, which can address the challenge of insufficient and imbalanced training data,
and the training time is negligible compared to neural network based approaches.

\subsection{Classification Model} 
In our machine learning framework, we address the challenges of using real-world PMU data through data preprocessing, fine-grained event extraction, and feature engineering. The proposed machine learning framework can significantly reduce the dimensionality of the training data to a few number of good features, which can facilitate the use of many off-the-shelf lightweight models.
To train an event classification model, we use the constructed features as input and the corresponding event type in the event logs as label. 
To find the best model, we train different machine learning models and compare their performance.
In our experiments (see Section \ref{sec:exp}), we examine the performance of many off-the-shelf models, in which the Random Forest model performs best.

\subsection{Event Classification Using Real-world PMU Data}

\begin{figure}[!ht]
\centering
\includegraphics[scale=0.42]{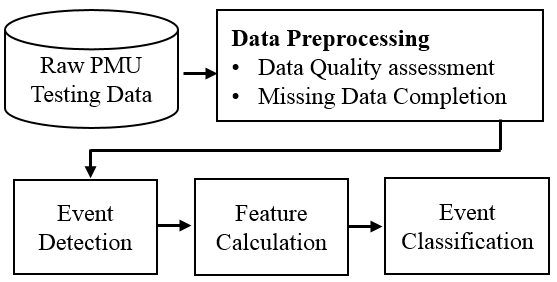}
\caption{The workflow for event classification using real-world PMU data.}
\label{fig:test}
\end{figure}

To apply the trained event classifier in practice, we still need to deal with the incomplete and noisy PMU measurements, which are streaming into the system in real time. Based on the proposed machine learning framework, we develop a workflow for event classification using real-world PMU data, as illustrated in Fig. \ref{fig:test}. Specifically, the raw PMU data are streaming into the system. The data preprocessing (i.e., data quality assessment and missing data completion) developed in Section \ref{sec:preprocessing} is first applied on the raw PMU data to fix the bad and missing data. Then, the event detection is used to determine whether there is an event. If an event occurs, the PMU data in the small window $W$ will be used to calculate the features based on Section \ref{sec:feature}. The calculated features will be input into the event classifier to determine the event type.




\section{Case study of real-world PMU data}\label{sec:exp}
\subsection{Experimental Setup} 
\subsubsection{Data} 
The data used in case studies are described in Section \ref{sec:data}. 
For the 2-year PMU data, we randomly split the PMU data in each month into 80\% for training and 20\% for testing, so that the events across the entire year are fairly distributed between the training set and the testing set.
The event types provided in the event logs are used as the labels for classification, except the line trip  and the oscillation events, because the recorded line trip cannot be determined as faults based on the current high-level description of the event logs and oscillation events can be detected/classified by many existing approaches (e.g., \cite{nabavi2015distributed,ma2021application}), which can be easily integrated into the proposed machine learning framework. Similar to the related work \cite{yuan2021learning}, this paper focuses on the classification of line outages, transformer outages, and frequency events.

\subsubsection{Evaluation Metrics} 
Four metrics are used to evaluate the classification performance, i.e., accuracy (ACC), precision (PRE), recall (REC), and F1 score, which  are defined as follows:
\begin{align}
\text{ACC} &= \text{(TP+FN)/(TP+TN+FP+FN), }\nonumber\\
\text{PRE} &= \text{TP/(TP + FP), }\nonumber\\
\text{REC} &= \text{TP/(TP + FN), }\nonumber\\
\text{F1} &= 2 \times \text{(PRE} \times \text{REC)/(TP + FP), }
\nonumber
\end{align}
where TP (i.e., True Positive) and TN (i.e., True
Negative) denote the number of positive and negative instances that
are correctly classified, respectively. FP (i.e., False Positive) and FN (i.e., False Negative) denote the number of misclassified negative and
positive instances, respectively. 
The precision
represents how good the proposed model is at excluding false
alarms. The recall represents how good the
proposed model is at not missing true events.

\subsubsection{Parameter Tuning} 
\begin{figure}[!t]
\centering
\includegraphics[scale=0.4]{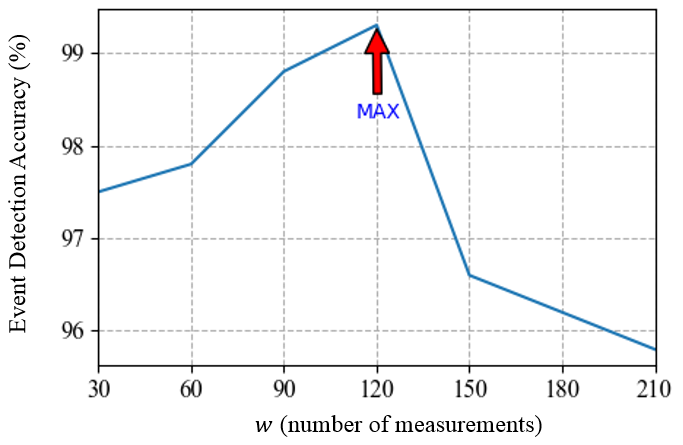}
\caption{Event detection accuracy versus $w$.
}
\label{fig:wsize}
\end{figure}

In the proposed machine learning framework, there are two key parameters, i.e., the number of measurements $w$ for event detection and the window size $W$ for feature calculation. Based on Algorithm \ref{alg:bo}, we find the best $w$ is 120 (detection accuracy: 99.3\%), which is used throughout the experiments. To show the impact of $w$ on the event detection, Fig. \ref{fig:wsize} illustrates the event detection accuracy obtained under different $w$.


\begin{figure}[!t]
\centering
\includegraphics[scale=0.4]{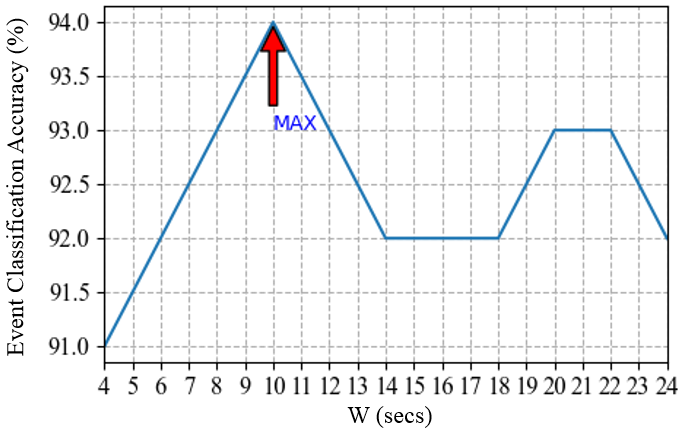}
\caption{Event classification accuracy versus $W$.
}
\label{fig:rlength}
\end{figure}

For the window size $W$, we optimize $W$ based on the classification accuracy. In Fig. \ref{fig:rlength}, we evaluate the classification accuracy under different $W$ and find the best accuracy is achieved when $W$ is 10 seconds, where the Random Forest model is used.


\subsubsection{Benchmark} 
We evaluate the performance of different classification models under the proposed machine learning framework:
\begin{itemize}
    \item The Random Forest (RF) model,
    \item The Gradient Boosting Decision Tree (GBDT) model,
    \item The K-Nearest Neighbor (KNN) model,
    \item The Logistic Regression (LR) model,
    \item The Support Vector Machine (SVM) model,
    \item The Decision Tree (DT) model.
\end{itemize}
We use the radial basis function (RBF) as the kernel of SVM. We fine-tune the hyperparameters of all these models via grid search.
We also compare the performance of these models with neural network models, i.e., the Long Short-Term Memory (LSTM) model and the Convolutional Neural Networks (CNN) model, where the features are automatically generated based on the input signals. Specifically, for LSTM and CNN, 3-minute measurements of each event (1 minute before and 2 minutes after the start timestamp of a detected event) are used as input to train the classifiers. When training these models, 5-fold cross-validation is used. The trained models are then evaluated using the testing dataset.

\subsection{Classification Results}


\begin{table}[H]
  \centering
  \captionsetup{justification=centering}
  \caption{event classification performance under different models.}
  \label{tab:res_compare}
    \scalebox{1}{
\begin{tabular}{c|cccc}
\hline
Model  & ACC(\%) &PRE(\%) &REC(\%) &F1(\%) \\
\hline
RF & \textbf{94} & \textbf{95}  & \textbf{88}  & \textbf{91} 
\\
GBDT & 93 & 94  & \textbf{88}   & \textbf{91} 
\\
KNN & 85 & 78  & 63  & 66
\\
LR & 93 & 81  & 66  & 69
\\
DT & 81 & 78  & 81  & 79
\\
SVM & 92 & 86  & 74  & 79 
\\
CNN & 69 & 23  & 33  & 27 
\\
LSTM & 69 & 23  & 33  & 27 
\\
\hline
\end{tabular}}
\end{table}


Table \ref{tab:res_compare} compares the performance of different classification models using the testing data. The performance of the RF model outperforms the other models in all evaluation metrics. From Table \ref{tab:res_compare}, we can observe that the performance of non-neural network models (i.e., RF, GBDT, KNN, LR, SVM, DT) trained using the proposed machine learning framework is better than the neural network models (i.e., CNN, LSTM) trained directly using the PMU data. 
For the non-neural network models, the performance of RF and GBDT are close, and both performs much better than KNN, LR, SVM, and DT. Compared to KNN, LR, SVM, and DT, RF and GBDT are ensemble methods, which consist of a pool of trees that can better capture the features of different types of events and deal with the imbalanced training data. 



\begin{figure}[!t]
\centering
\includegraphics[scale=0.44]{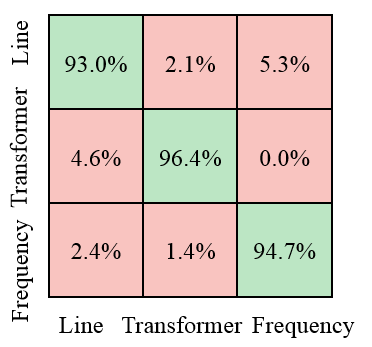}
\caption{Confusion matrix of the RF model for the testing data.}
\label{fig:res}
\end{figure}

Fig. \ref{fig:res} illustrates the confusion matrix of the RF model based on the testing data, where the rows represent the estimated event type, the columns represent the true event type, and the value of each cell represents the classification accuracy of the corresponding event type. The diagonal and off-diagonal cells in Fig. \ref{fig:res} represent the events that are correctly and incorrectly classified, respectively.
From Fig. \ref{fig:res}, it is observed that most of the events (i.e., 94\% on average) can be classified correctly. 
For the misclassified events, 4.6\% of line outages are misclassified as transformer outages, and 5.3\% of frequency events are misclassified as line outages. By analyzing these misclassified events, we find that the reasons for misclassifying these events are 
1) the patterns of some line and transformer outages are similar and 2) the patterns in the ROCOF signal of some frequency events are buried by noise due to the low SNR.

For the neural network models, the classification results in Table \ref{tab:res_compare} agree with the findings in \cite{yuan2021learning,shi2021power} that directly using the PMU data to train neural networks cannot achieve good classification results. This is because 1) directly applying such imbalanced data can cause severe overfitting problems; 2) the number of labeled events is only a few thousand\revv{s}, which is not enough for training a good neural network; and 3) neural network based automatic feature generation using the PMU data cannot well capture the event patterns. Hence, the performance of off-the-shelf neural network models directly trained using such a dataset can be significantly poor.



\begin{table}[H]
  \centering
  \captionsetup{justification=centering}
  \caption{Training time of different non-neural network models. }
  \label{tab:trainT}
    \scalebox{1}{
\begin{tabular}{c|cccccc}
\hline
Model & RF &GBDT &KNN &LR &DT &SVM \\
\hline
Training time(secs) &1.59 &25.39 &0.01  &0.37  &0.21 &3.48 
\\
\hline
\end{tabular}}
\end{table}

Table \ref{tab:trainT} compares the training time of different non-neural network models in a server with
dual-sockets Intel(R) Xeon(R) CPU E5-2630 v4 @ 2.20GHz and 64 GB of memory. The training of all these models takes only a few seconds, compared with hours of training time for neural networks reported in \cite{yuan2021learning,shi2021power}. Specifically,  the RF model, which achieves the best classification performance, takes only 1.59 seconds for training. In comparison, the training time of the GBDT model is about 25 seconds. These results show that the proposed machine learning framework provides a promising ways to train good event classifiers with good interpretability and low training cost.


\subsection{Robustness Analysis} 

Event classifiers trained under the proposed machine learning framework are robust against the low-quality PMU data (e.g., bad and missing data described in Section \ref{sec:preprocessing}). To evaluate the robustness, we compare the classification performance under different data missing rates in the testing dataset. Specifically, the PMU data are assumed to be missing randomly with a probability. The missing data will be first recovered in our framework using our regularized tensor completion approach \cite{ghasemkhani2020regularized}. Then, the recovered PMU data will be used to calculate the features for event classification.  

\begin{table}[H]
  \centering
  \captionsetup{justification=centering}
  \caption{Event classification performance of the RF model under different missing rates}
  \label{tab:res_miss}
    \scalebox{1}{
\begin{tabular}{c|cccc}

\hline
Missing Rate(\%)  & ACC(\%) &PRE(\%) &REC(\%) &F1(\%)\\
\hline

10 & 94 & 94  & 88  & 91
\\
20 & 87 & 93  & 87  & 85

\\
30 & 87 & 89  & 63  & 69
\\
40 & 88 & 91  & 65  & 71
\\
50 & 87 & 91  & 64  & 69
\\

\hline
\end{tabular}}
\end{table}

Table \ref{tab:res_miss} compares the classification performance using the RF model under different missing rates. It is observed that the classification performance under 10\% missing rate is almost the same as the results in Table \ref{tab:res_compare}. As the missing rate increases, the classification performance slightly drops but still remains at a high level, while the classification performance of the neural network model in \cite{yuan2021learning} drops significantly.
This is because the proposed machine learning framework leverages the features constructed based on the event patterns and does not depend on the specific PMU measurements for event classification. Therefore, even with high missing rates, our approach can still perform well as long as the patterns are preserved.




\section{Conclusion}
In this paper, we have developed a machine learning framework for training robust event classifiers to enhance the situational awareness of power systems using PMU data. Our framework has addressed the challenges of using real-world PMU data, such as incomplete and noisy PMU measurements, inaccurate event timestamps in the event logs, and insufficient and imbalanced training data. 
One salient merit of the proposed framework is that large-scale real-world PMU data is reduced to a small set of event features, which can be used to efficiently train many off-the-shelf lightweight machine learning models. As the features are constructed based on the event patterns, the contribution of any single PMU measurement to the features is low, which can effectively mitigate the impact of bad and missing data and improve the robustness of the event classifiers. 
Numerical experiments using the real-world dataset from the Western Interconnection of the U.S power transmission grid show that the event classifiers trained under the proposed framework can achieve high classification accuracy while being robust against low-quality data. 
In conclusion, the proposed machine learning framework provides a promising way to train robust event classifiers with good interpretability and low training cost. 

\section*{Acknowledgment and disclaimer}

We thank Pacific Northwest National Laboratory (PNNL) for offering the data in this study.
This material is based upon work supported by the Department of Energy National Energy Technology Laboratory under Award Number DE-OE0000911. This work was prepared as an account of work sponsored by an agency of the United States Government. Neither the United States Government nor any agency thereof, nor any of their employees, makes any warranty, express or implied, or assumes any legal liability or responsibility for the accuracy, completeness, or usefulness of any information, apparatus, product, or process disclosed, or represents that its use would not infringe privately owned rights. Reference herein to any specific commercial product, process, or service by trade name, trademark, manufacturer, or otherwise does not necessarily constitute or imply its endorsement, recommendation, or favoring by the United States Government or any agency thereof. The views and opinions of authors expressed herein do not necessarily state or reflect those of the United States Government or any agency thereof.

\bibliographystyle{ieeetr}
\bibliography{ref}

\end{document}